\documentclass[letterpaper]{article} 
\usepackage{aaai25}  
\usepackage{times}  
\usepackage{helvet}  
\usepackage{courier}  
\usepackage[hyphens]{url}  
\usepackage{graphicx} 
\usepackage{natbib}  
\usepackage{caption} 
\usepackage{algorithm}
\usepackage{algorithmic} 

\usepackage{newfloat}
\usepackage{listings}
\DeclareCaptionStyle{ruled}{labelfont=normalfont,labelsep=colon,strut=off} 
\floatstyle{ruled}
\newfloat{listing}{tb}{lst}{}
\floatname{listing}{Listing}
\usepackage[dvipsnames]{xcolor}
\usepackage{booktabs}
\usepackage{amsmath}
\usepackage{amsfonts}
\usepackage{multirow}
\usepackage{lipsum}
\usepackage{makecell}
\usepackage{enumitem} 
\usepackage{pdfpages}
\usepackage{nccmath}
\usepackage{subcaption}
\usepackage{bm} 

\def\onedot{\ifx\@let@token.\else.\null\fi\xspace}

\newcommand{\Tref}[1]{Table~\ref{#1}}
\newcommand{\Eref}[1]{Eq.~(\ref{#1})}
\newcommand{\Fref}[1]{Fig.~\ref{#1}}

\newcommand{\Abref}[1]{Ablation.~\ref{#1}}
\newcommand{\model}{\text{ViPCap}}
\newcommand{\ourmodule}{\text{ViP}}

\title{ViPCap: Retrieval Text-Based Visual Prompts \\ for Lightweight Image Captioning}

\author{Taewhan Kim,
        Soeun Lee,
        Si-Woo Kim,
        Dong-Jin Kim\thanks{Corresponding Author.} \\
        }

\affiliations{
        \textsuperscript{}
        Hanyang University, South Korea \\
        \textsuperscript{}
        \{taewhan, soeun, boreng0817, djdkim\}@hanyang.ac.kr
        }

\begin{document}

\maketitle

\begin{abstract}
Recent lightweight image captioning models using retrieved data mainly focus on text prompts. However, previous works only utilize the retrieved text as text prompts, and the visual information relies only on the CLIP visual embedding. 
Because of this issue, there is a limitation that the image descriptions inherent in the prompt are not sufficiently reflected in the visual embedding space.
To tackle this issue, we propose $\model$, a novel retrieval text-based visual prompt for lightweight image captioning. $\model$ leverages the retrieved text with image information as visual prompts to enhance the ability of the model to capture relevant visual information. By mapping text prompts into the CLIP space and generating multiple randomized Gaussian distributions, our method leverages sampling to explore randomly augmented distributions and effectively retrieves the semantic features that contain image information.
These retrieved features are integrated into the image and designated as the visual prompt, leading to performance improvements on the datasets such as COCO, Flickr30k, and NoCaps. Experimental results demonstrate that $\model$ significantly outperforms prior lightweight captioning models in efficiency and effectiveness, demonstrating the potential for a plug-and-play solution. The source code is available at \url{https://github.com/taewhankim/VIPCAP}.
\end{abstract}

%


\section{Introduction}\label{sec:introduction}

\begin{figure}[t!]
    \centering
    \begin{subfigure}{0.45\textwidth}
          \centering
      \includegraphics[width=\textwidth]{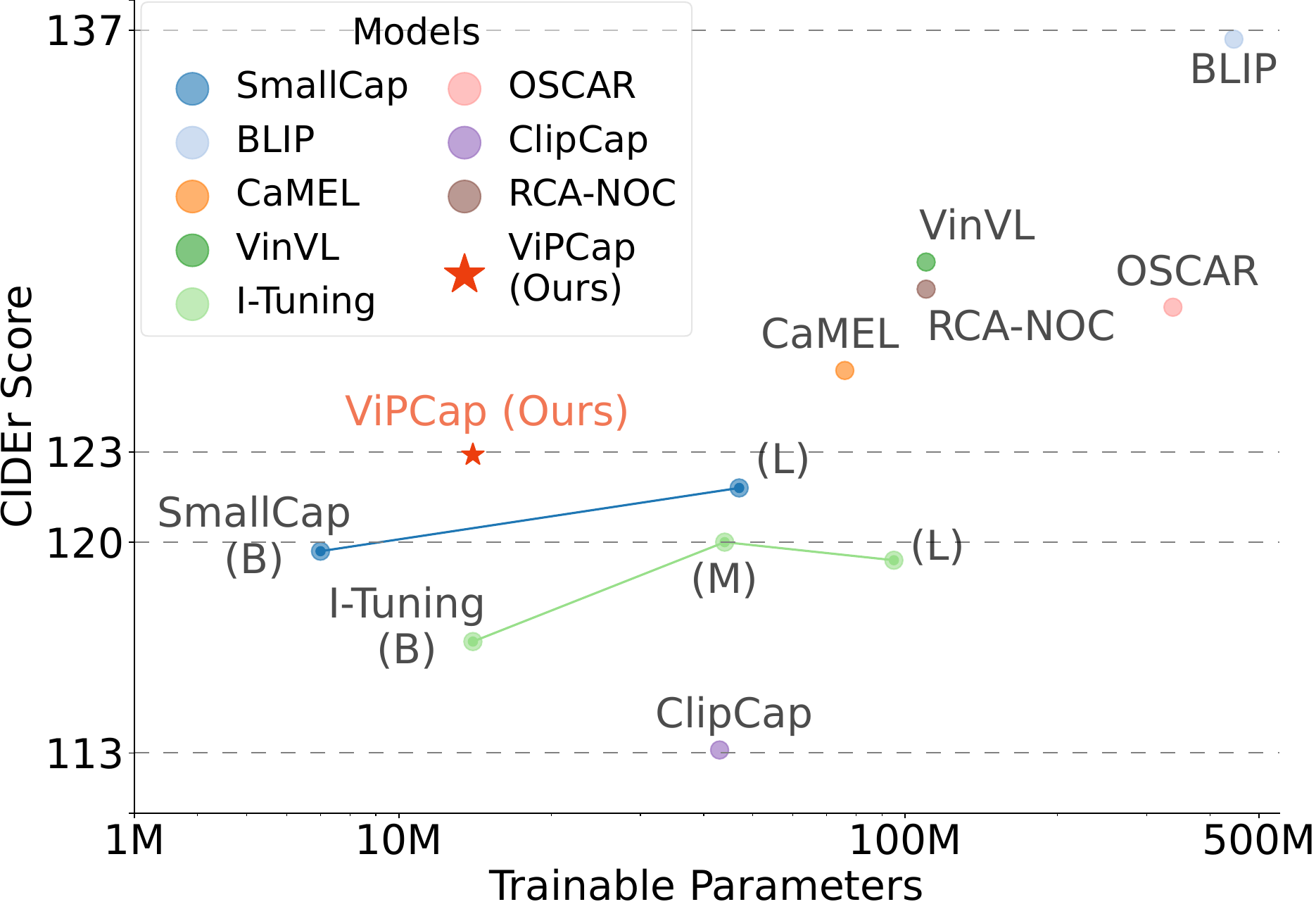}
      \caption{Comparison of trainable parameter sizes and CIDEr scores for each model. B, M, and L denote the Base, Medium, and Large models, respectively.}
    \label{fig:total_model_param}
    \end{subfigure}

    \begin{subfigure}{0.45\textwidth}
          \centering
      \includegraphics[width=\linewidth]{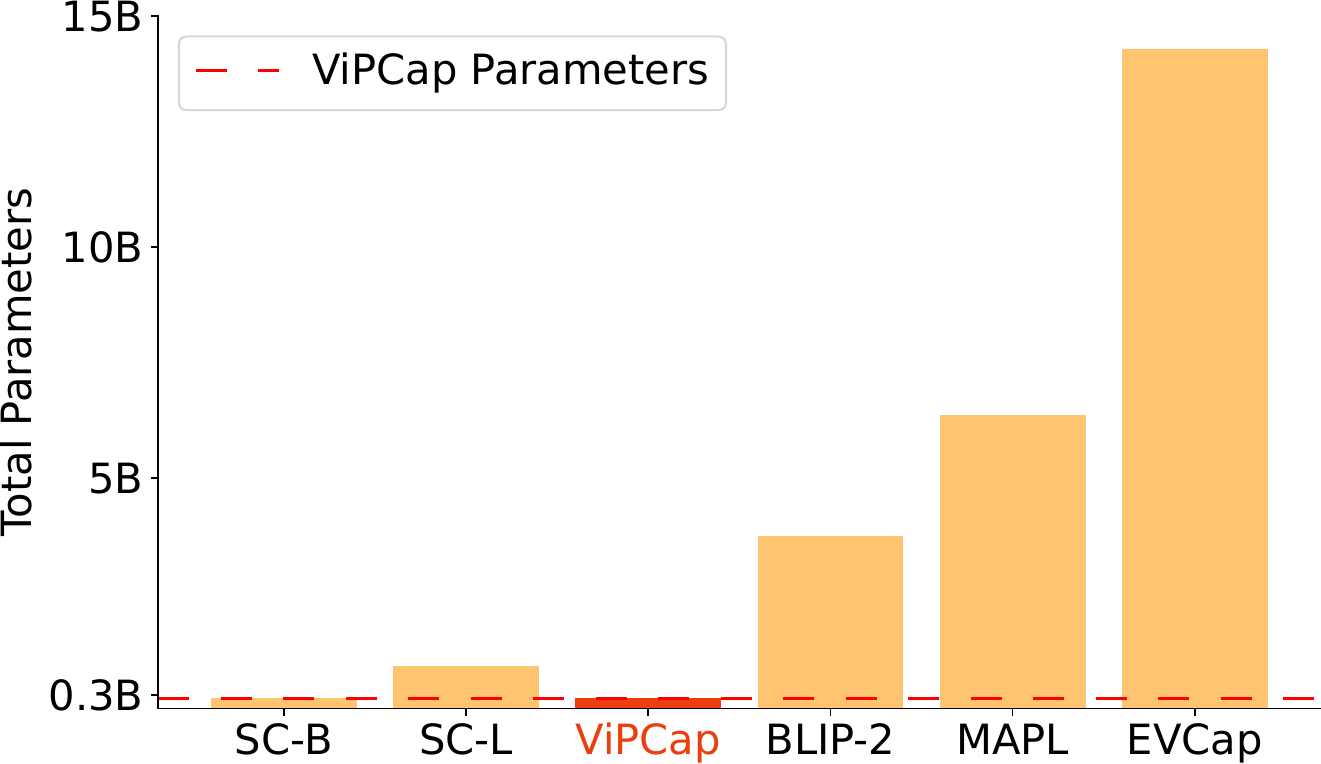}
      \caption{Comparison of total parameters between $\model$ (Ours) and other models. SC-B denotes SmallCap$_\text{Base}$ model, SC-L represents the SmallCap$_\text{Large}$ model.}
      \label{fig:total_model_param2}
    \end{subfigure}
 \caption{(a) $\model$ shows the best efficiency among the lightweight captioning models. (b) EVCap and MAPL require large models such as EVA-CLIP~\cite{evaclip}, Vicuna~\cite{vicuna2023}, and GPT-J~\cite{gpt-j}. With under 0.3B total parameters, $\model$ achieves competitive performance despite its small size.
}
\label{fig:fig_all_parameters}
\end{figure}

Vision and language (V\&L) tasks, such as image captioning, have advanced with large-scale models like SimVLM (1.4B), PaLi (3B), and REVEAL (2.1B)~\cite{simvlm, pali, reveal}. However, these advanced multimodal models require a large number of parameters, resulting in high computational and dataset construction costs.

To enhance training efficiency, recent works suggest focusing on learning a mapping network, such as MAPL and BLIP-2~\cite{mapl, blip2}, that bridges the modality gap between images and text or training learnable tokens like EVCap~\cite{evcap}.
For example, BLIP-2 introduces the Qformer, which aligns the two modalities while keeping a V\&L models frozen. 
Nevertheless, despite the efficiency of this approach, BLIP-2 still requires over 1B trainable parameters. Although EVCap trains learnable tokens with few trainable parameters, EVCap and MAPL depend on high-performance models like EVA-CLIP~\cite{evaclip} and Vicuna~\cite{vicuna2023}, with over 5B total parameters as shown in \Fref{fig:total_model_param2}.

\begin{figure}[t]
\centering
  \includegraphics[width=0.9\linewidth]{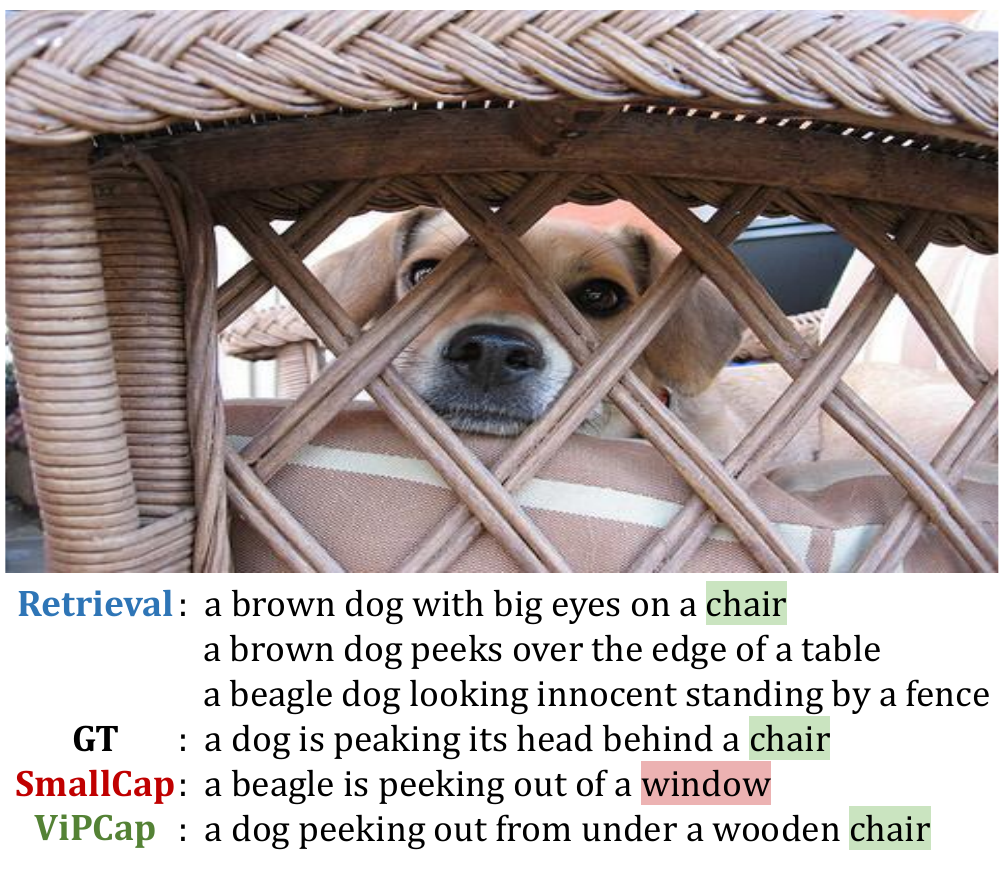}
  \caption{SmallCap~\cite{smallcap} fails to accurately represent visual information, such as a local object, in the ground truth (GT) or retrieval text (Retrieval). 
  In contrast, our $\model$ accurately captures visual information found in GT or retrieval text.}
  \label{fig:failure}
\end{figure}

Recent models like SmallCap, EXTRA, and Re-ViLM~\cite{smallcap, extra, revilm} are emerging to reduce computational costs using external knowledge. These models retrieve texts semantically similar to input images and use them as only text prompts. However, there is a limitation in that visual information relies only on the CLIP vision encoder.
As depicted in \Fref{fig:failure}, SmallCap, which uses retrieval caption as a text prompt without visual prompts, encounters an issue where it can not contain detailed visual descriptions in a caption. We suspect this is because the image description in text prompt is not utilized as visual information.

In this paper, we propose a retrieval text-based visual prompt for lightweight image captioning (\textbf{$\model$}), leveraging retrieved texts describing image information as \emph{visual prompts}.
First, given that the retrieved text provides a comprehensive image description, we encode the text prompt into the CLIP embedding space and transform it into patch-level hidden representations to extract semantic information. 
To effectively enhance local visual representations using a global text representation, we randomly augment the semantic representation from the text prompt.
In particular, we assume that the embedding vector follows a randomized Gaussian distribution and extract $M$ semantic features from this distribution as the basis for a visual prompt.

Previous works, including CapDec and LinCIR~\cite{capdec, lincir}, address the modality gap in V\&L tasks using Gaussian distributions. Based on this approach, we propose modeling text features following a randomized Gaussian distribution. Unlike the heuristic approaches like CapDec, our approach generates semantic features sampled from a learnable distribution, aiming for a high correlation with visual features. We assume these semantic features contain visual information and expect them to closely resemble the input image features. 
To achieve this, we employ a patch retrieval module that aligns semantic features with each input image patch.

The retrieved patch features are combined with the input image features to generate the
\emph{visual prompt} is added with the image features before decoder input. This approach aims to enhance the model's ability to capture relevant visual representations.
To the best of our knowledge, our work is the first to utilize retrieval text as a visual prompt for lightweight image captioning.

Our approach achieves superior performance
on the COCO dataset~\cite{mscoco} compared to our baseline model, SmallCap, and it significantly improves performance over previous lightweight models on the NoCaps dataset~\cite{nocaps}.
In the experiments, we integrate our $\ourmodule$ module into retrieval-based models, text-only training models, and various prompts, resulting in consistent performance improvements.
 
Our contribution can be summarized as follows:
(1) We propose a novel visual prompt for lightweight image captioning models named $\model$, which leverages retrieved texts to generate visual prompts.
(2) We introduce the $\ourmodule$ module, which retrieves semantic information from text features and combines it with image features to generate the visual prompt.
(3) Extensive experiments demonstrate that our method is efficient and outperforms previous models across datasets like COCO and NoCaps, regardless of the text prompt types used.

\begin{figure*}[t!]
  \centering   \includegraphics[width=\textwidth,height=\textheight,keepaspectratio]{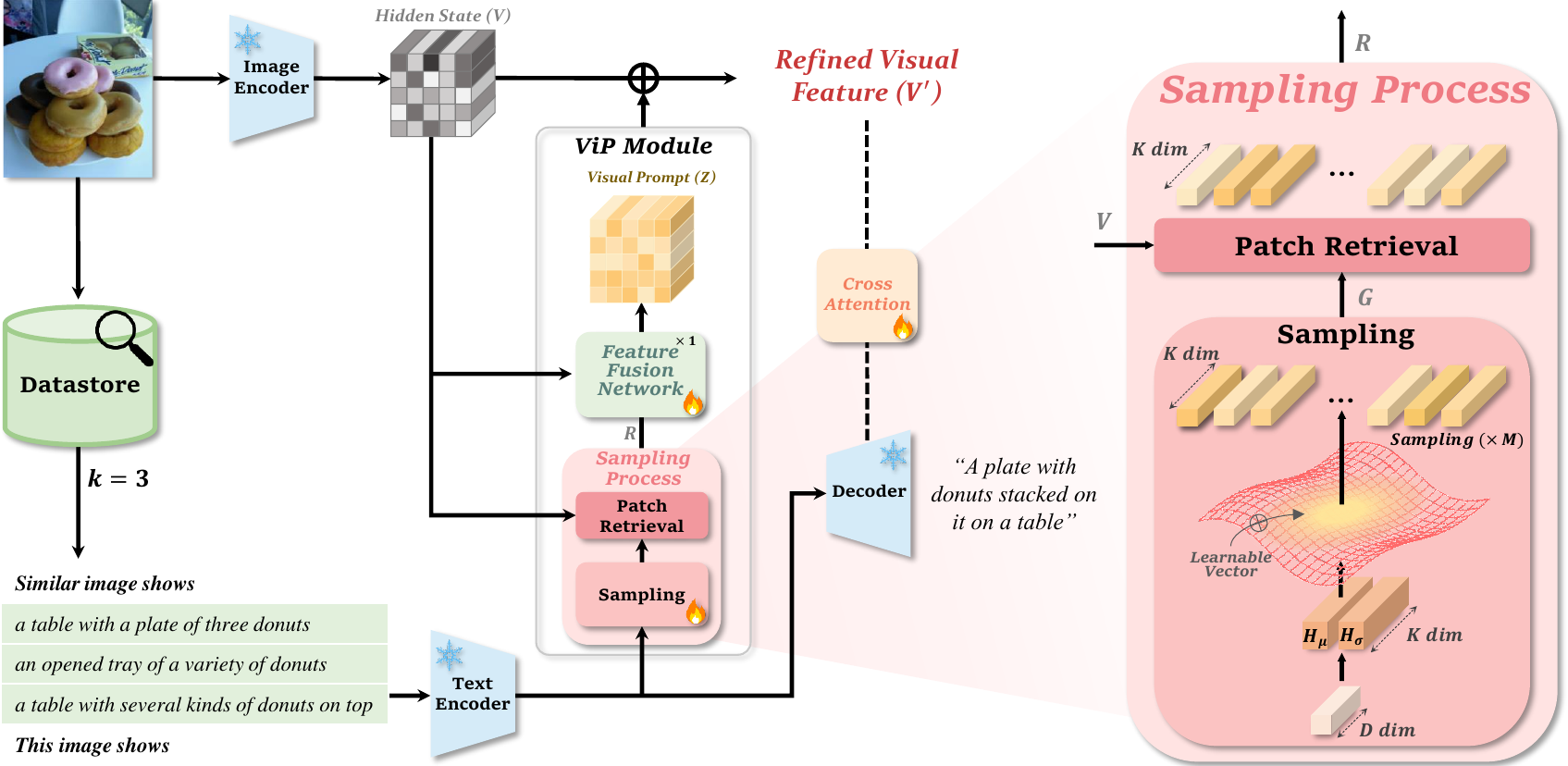}
   \caption{
$\model$ leverages the CLIP text encoder to extract retrieved text features for visual prompts generation. 
The $\ourmodule$ module performs $M$ sampling iterations from the text embedding distribution to extract semantic features $G$ closely aligned with image patch features $V$. 
Then, in the patch retrieval module, we retrieve semantic vectors from $G$ that are highly correlated with image patch features $V$.
The retrieved semantic features $R$ are fused with image features $V$ within the Feature Fusion Network, and the resulting output is set as the visual prompt $Z$. Finally, the refined visual feature $V'$ via summation with the visual prompt is fed to the decoder through the cross-attention layer.
   }
  \label{fig:VipCap_framework}
\end{figure*}


\section{Related Work}
\noindent\textbf{CLIP in captioning.}
With the advent of vision-language models ~\cite{clip, ALIGN}, significant advancements have been made in V\&L tasks. Notably, CLIP achieves multimodal alignment with 400M image-text pairs. The image captioning models that use a clip-based encoder such as BLIP-2, LLaVA, and  MiniGPT-4~\cite{blip2, llava, minigpt4} create a mapping network and pass input features to the decoder. In our study, we map the CLIP text encoder to the randomized Gaussian distribution, desiring to achieve a high correlation with visual features and leveraging it to generate visual prompts, 

\noindent\textbf{Prompt tuning.}
Initially used in NLP~\cite{power, prefixtuning}, this approach has extended to V$\&$L and vision-only models~\cite{vpt, dualprompt} to utilize pretrained knowledge for downstream without parameter training. 
Recent techniques focus on trainable tokens instead of handcrafted prompts~\cite{handmaking}. Visual prompt tuning~\cite{vpt} adds learnable tokens to ViT for downstream tasks, while some methods~\cite{exploring} integrate task-specific patches at the pixel level. Although previous methods achieved some success, they still face challenges with complex tasks like captioning. We introduce a visual prompt method suitable for captioning.

\noindent\textbf{Retrieval for captioning.}
External knowledge helps reduce the cost of creating large image-text datasets for Multimodal LLM. Methods like SmallCap~\cite{smallcap}, LMCap~\cite{lmcap}, Knight~\cite{knight}, and MeaCap~\cite{meacap} use text-based datasets and store captions in a datastore. SmallCap~\cite{smallcap}, for example, generates captions using an input image and related text from a datastore with only 7 million trainable parameters, allowing fast and easy training, although it lacks visual knowledge. Our work enhances performance by generating visual prompts from retrieved text with detailed image
descriptions.


\section{Proposed Method}

Our model adopts SmallCap~\cite{smallcap} as a baseline, which retrieves enriched expressions from an external datastore and integrates the pretrained CLIP encoder~\cite{clip} with GPT-2~\cite{gpt2} through cross-attention layers. 
SmallCap consists of the retrieved texts with the following hard prompt and inputs it into the decoder. Similarly, we utilize this approach in encoding our prompts: \texttt{Similar images show} \{\texttt{caption$_1$}\}\texttt{...}\{\texttt{caption$_k$\}}. \texttt{This image shows \_\_\_}. However, SmallCap primarily focuses on text prompts, while visual representation relies solely on the performance of the vision encoder. 

In this work, we introduce $\model$, a novel approach that enhances the performance of lightweight image captioning by generating \emph{visual prompts} based on the semantic information in text prompts as illustrated in \Fref{fig:VipCap_framework}.
$\ourmodule$ module encodes retrieved texts into the CLIP embedding space and converts them into patch-level representations.
Then, assuming a multivariate Gaussian distribution, our method generates semantic features $M$ times 
to obtain semantic features that are highly correlated to the local visual features.
Unlike the heuristic Gaussian distribution approach in Capdec~\cite{capdec},
our novel approach creates semantic features through a learnable distribution. Then, the semantic features are leveraged by the patch retrieval module to closely align with the input image patches. The matched semantic features are fused with the input image to generate the visual prompt. Finally, we train cross-attention layers to fuse the features in the decoder.

\subsection{Randomized Gaussian Distribution Sampling}\label{sec:sampling}
The $\ourmodule$ module aims to effectively sample the semantic information from the retrieved texts to enrich the input visual feature with semantic content in the form of visual prompts. Given retrieved texts $T$, our model encodes the retrieved text into $D$ dimensional vector using pretrained CLIP text encoder $\phi(\cdot)$. 
Also, the CLIP image encoder embeds the input image into $K$ dimensional visual features $V = \{\vec{v}_1, \vec{v}_2, \ldots, \vec{v}_N\} \in \mathbb{R}^{N\times K}$ representing $N$ number of patch-level visual features.

When generating visual prompts, a single text feature might be insufficient to provide the necessary details to generate a visual prompt with complex patch-level local information.
To address this, we employ 
a random
augmentation techniques to sample semantic features from the Gaussian distribution.
Also, instead of learning multiple mapping functions for local regions individually, we empirically find that sampling random vectors helps better match with visual local representations.

We estimate the parameters of distribution of the text embedding $\vec{\mu},\vec{\sigma} \in \mathbb{R}^{K}$ assuming it follows a multivariate Gaussian distribution $\mathcal{N}(\vec{\mu}, \vec{\sigma}^2 I$).
We design functions $\mathcal{H}_{\mu}(\cdot)$ and $\mathcal{H}_{\sigma}(\cdot)$ to map text features into the mean and standard deviation of multivariate Gaussian distribution. These functions are implemented as MLP layers to map from $D$ dimensions to $K$ dimensions while sampling the mean and standard deviation  
($\mathcal{H}:\mathbb{R}^{D}\rightarrow \mathbb{R}^{K}$).
We empirically find that adding an additional learnable vector $\vec{\omega_{add}}$ with a hyperparameter $\alpha$ as a scaling factor to the MLP shows better performance and captures complex data structures more effectively. The $\alpha$ is used to expand the range of the learnable vector.
Let $\vec{\mu}$ and $\vec{\sigma}$ are computed via $\vec{\mu} = \mathcal{H}_{\mu}(\phi(T))+ \alpha \cdot \vec{\omega_{add}}$, and  $\vec{\sigma} = \mathcal{H}_{\sigma}(\phi(T))$, respectively.

$\ourmodule$ module samples $M$ number of semantic features from this learnable Gaussian distribution to obtain semantic features that are highly correlated to the local visual embedding.
We define the set of semantic representation 
${G} \in \mathbb{R}^{M \times K}$ obtained from the text features as:
\begin{align}
    G & = \{ \vec{g}_i\sim \mathcal{N}(\vec{\mu},\vec{\sigma}^2 I ; \phi(T))\}^M_{i=1}.
\end{align}
Additionally, increasing $M$ allows semantic features ${G}$ to deliver better fine-grained visual information to the input feature (See more details in \Abref{sec:M_num}).
Through the reparameterization trick~\cite{vae}, $\vec{g}$ can be re-formulated as $\vec{g} = \vec{\mu} + \vec{\sigma} \cdot \vec{\epsilon},$ where $\vec{\epsilon} \sim {N}(0,I)$.

\subsection{Patch Retrieval Module for Semantic Features}

\begin{figure}[t]
  \includegraphics[width=\linewidth]{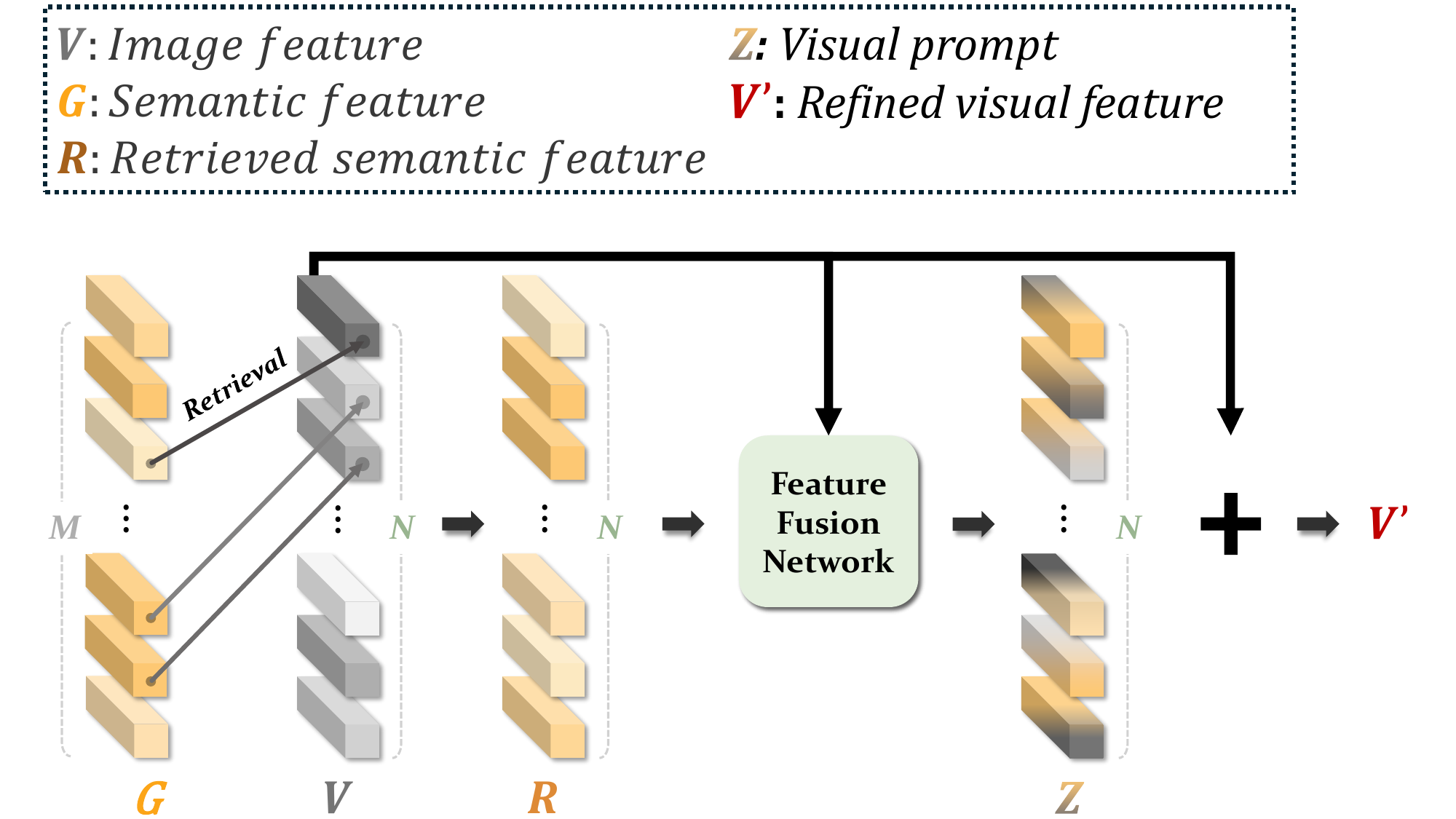}
  \caption{Calculating similarity between input feature $V$ and the semantic features $G$. $V$ retrieves essential semantic representations from $G$ and combines these retrieved semantic features $R$ through the fusion network. The fusion network generates a visual prompt $Z$ by integrating image features $V$. After that, the visual prompt and image features are combined to create refined visual features $V'$.}
  \label{fig:patch_retrieval}
\end{figure}

We hypothesize that the semantic features $G = \{\vec{g}_1, \vec{g}_2, \ldots, \vec{g}_M\}$ sampled from the Gaussian distribution contain the textual information describing the image.
To effectively leverage random semantic features, we retrieve semantic features that contain useful visual information related to visual prompts based on feature similarity.
In particular,
we introduce a patch retrieval module, as depicted in \Fref{fig:patch_retrieval}, that compares the similarity between image patch-level representations $V$ and the semantic features $G$. This module employs cosine similarity, denoted as $sim(\cdot,\cdot)$, to effectively identify the most relevant semantic information for each patch representation $\vec{v}_i\in V$. Consequently, $\model$ chooses $N$ relevant vectors from $M$ candidates, one responsible for each patch vector, with high similarity to the input feature, and generates $R$:

\begin{equation}
\begin{aligned}
& R = \{\vec{g}_{\mathcal{I}(j)}\}_{j=1}^{N} \in \mathbb{R}^{N \times K}, \\
& \quad \text{where } \mathcal{I}(j)={\mathrm{argmax}_{i\in [1:M]} }sim(\vec{g}_i, \vec{v}_j).
\end{aligned}
\end{equation}
This simple calculation process extracts valuable information through $R$ without any additional training. The extracted vectors have the potential to provide semantic representations that the vision encoder cannot offer.

\subsection{Feature Fusion Network and Visual Prompts}

After obtaining the representations $R$ containing semantic information relevant to the visual features, we integrate them with the input image to enhance the visual feature to generate the \emph{visual prompt} $Z$.
We introduce a Feature Fusion Network (FFN) designed to effectively fuse the visual representations $V$ and the retrieved semantic features $R$, as shown in \Fref{fig:patch_retrieval}, by bridging the gap between them.

FFN is designed with a transformer architecture that incorporates both self-attention and cross-attention layers. Unlike previous mapping networks~\cite{blip2, clipcap} that typically consist of 8 to 12 layers, this module uses only a single layer ($l=1$). This efficiency results from earlier stages where features are sampled to closely align with visual representations, making a single layer sufficient for effective fusion. 
This network uses the input feature $V$ as the query and the retrieved semantic representation $R$ as the key.

Note that a simple summation between $V$ and $R$ ignores distribution differences between features, and a simple concatenation cannot consider relational context between modalities (see \Abref{sec:ffn_design} for more details).


\begin{table*}[t]
\begin{center}

\begin{tabular}{l|c|cccc|cc|cccc}
\toprule[1pt]
\multirow{3}{*}{\textbf{Method}} & \multicolumn{1}{c|}{\textbf{Training}} & \multicolumn{4}{c|}{\textbf{COCO}} & \multicolumn{2}{c|}{\textbf{Flickr30k}} & \multicolumn{4}{c}{\textbf{NoCaps}} \\
& \multicolumn{1}{c|}{\textbf{Param}} & \multicolumn{4}{c|}{\textbf{Test}} & \multicolumn{2}{c|}{\textbf{Test}} & \multicolumn{4}{c}{\textbf{Val}}  \\
& \multicolumn{1}{c|}{$\theta$} & \multicolumn{1}{c}{B@4} & \multicolumn{1}{c}{M} & \multicolumn{1}{c}{C} & \multicolumn{1}{c|}{S} & \multicolumn{1}{c}{C} & \multicolumn{1}{c|}{S} & \multicolumn{1}{c|}{In} & \multicolumn{1}{c|}{Near} & \multicolumn{1}{c|}{Out} & \multicolumn{1}{c}{Entire} \\ 
\midrule
\multicolumn{12}{l}{\textbf{Large scale training models}} \\
OSCAR$_\text{Large}$~\shortcite{oscar} & 338M & 37.4 & 30.7 & 127.8 & 23.5 & - & - & 78.8  & 78.9  & 77.4  & 78.6  \\
LEMON$_\text{Huge}$~\shortcite{lemon} & 675M & 41.5 & 30.8 & 139.1 & 24.1 & - & - & 118.0  & 116.3  & 120.2  & 117.3  \\
SimVLM$_\text{Huge}$~\shortcite{simvlm} & 632M & 40.6 & 33.7 & 143.3 & 25.4 & - & - & 113.7  & 110.9  & 115.2  & 112.2  \\
BLIP2$_\text{ViT-g OPT$_\text{2.7B}$}$~\shortcite{blip2} & 1.1B & 43.7 & - & 145.8 & - & - & - & 123.0  & 117.8  & 123.4  & 119.7  \\ 
CogVLM~\shortcite{cogvlm} &1.5B& - & - & 148.7 & - & 94.9 & - & - & -  & 132.6  & 128.3  \\
PaLI$_\text{mT5-XXL}$~\shortcite{pali} &1.6B& - & - & 149.1 & - & - & - & -  & -  & -  & 127.0 \\
\midrule
\multicolumn{12}{l}{\textbf{Lightweight models}} \\
CaMEL~\shortcite{camel} & 76M & \textbf{39.1} & \textbf{29.4} & \textbf{125.7} & \textbf{22.2} & - & - & - & -  & -  & - \\
I-Tuning$_\text{Medium}$~\shortcite{ituning} & 44M & 35.5 & \underline{28.8} & 120.0 & \underline{22.0} & \textbf{72.3} & \textbf{19.0} & \underline{89.6}  & 77.4 & 58.8  & 75.4  \\
ClipCap~\shortcite{clipcap} & 43M & 33.5 & 27.5 & 113.1 & 21.1 & - & - & 84.9  & 66.8 & 49.1  & 65.8  \\
I-Tuning$_\text{Base}$~\shortcite{ituning} & 14M & 34.8 & 28.3 & 116.7 & 21.8 & 61.5 & 16.9 & 83.9  & 70.3  & 48.1  & 67.8  \\
SmallCap~\shortcite{smallcap} & 7M & 37.0 & 27.9 & 119.7 & 21.3 & 60.6 & - & 87.6 & \underline{78.6}  & \underline{68.9}  & \underline{77.9}  \\
SmallCap$_\text{d=16, Large}$~\shortcite{smallcap} & 47M & 37.2 & 28.3 & 121.8 & 21.5 & - & - & - & - & - & - \\
$\model$ (Ours) & 14M & \underline{37.7} & 28.6 & \underline{122.9} & 21.9 & \underline{66.8} & \underline{17.2} & \textbf{93.8} & \textbf{81.6} & \textbf{71.5} & \textbf{81.3}  \\
\bottomrule[1pt]
\end{tabular}
\caption{Comparison with large pre-trained and lightweight models with existing methods on the COCO test, Flickr30k test, and NoCaps val set.  CIDEr score is used for NoCaps evaluation. Our method shows the competitive performance in most metrics. }\label{tb:main}
\end{center}
\end{table*}

Finally, we obtain refined visual features $V'$ with a simple summation ($V' = V + Z$). 
Because the visual prompt closely aligns with the distribution of image features, we enable model to  generate the refined visual features through simple summation.
This method allows models like SmallCap, which have a vision encoder and language decoder, to use refined visual feature $V'$ instead of the input feature $V$, preserving the existing decoder design. This approach functions as a module that can be applied to various frameworks using encoders.

The decoder takes the input text embedding, while the refined visual feature $V'$ is included conditionally when computing the loss function:

\begin{equation}
L_\theta=-\sum_{i=1}^Q \log P_\theta\left(y_i \mid y_{<i}, V' ; \theta\right).\label{eq:loss}
\end{equation}
In \Eref{eq:loss}, the main difference in our approach is the use of refined visual feature $V'$ instead of $V$. In the cross-attention layers ($\theta$), weights are optimized by reducing the cross-entropy loss associated with forecasting the $Q$ tokens in the given reference $y_1, \ldots, y_q$.


\section{Experiments}

In this section, we conduct experiments to show the effectiveness of $\model$ over existing methods. First, we evaluate the advantages of our proposed model over the baseline in \Tref{tb:main} for in-domain and out-of-domain, and then we apply the $\ourmodule$ module with the text-only training model to identify the role of $\ourmodule$ module in \Tref{tb:Vipmodule_all_domain}, respectively. In \Tref{tb:noise_type}, $\ourmodule$ module evalutates performance of sampling method from different probability distributions. Moreover, we test the plug-and-play solution with different models in \Tref{tb:model_agnostic}, \Tref{tb:model_agnostic2}, and with different prompt styles in \Tref{tb:Vpp_prompt_agnostic}.

\subsection{Experimental Setup}

\noindent\textbf{Training dataset.}
We conduct experiments on image captioning benchmarks, i.e., COCO dataset ~\cite{mscoco}, NoCaps~\cite{nocaps}, Flickr30k~\cite{flickr30k}. For COCO and Flickr30k, we follow the Karpathy split~\cite{karpathy} used in the image captioning. We evaluate our model on the COCO and Flickr30k test set and NoCaps validation and test datasets, as well as the cross-domain experiments.

\noindent\textbf{Training setup.} $\model$ includes a vision encoder (ViT-B/32) and a language decoder (GPT-2$_\text{Base}$). Both are frozen during training. We train the $\ourmodule$ module and cross-attention layer. 
The FFN and cross-attention layer include a 12-head cross-attention layer with a single-layer block. 
To reduce the computational cost, we scale the dimension of cross-attention layers from 64 to 16.
$\model$ requires 14M training parameters and is trained on a single NVIDIA 6000 GPU with a batch size of 128.



\begin{table*}[t]
\begin{center}
{
\small

\setlength{\tabcolsep}{1mm} 
\begin{tabular}{l|cccc|cccc|cccc|cccc|cccc}
\toprule[1pt]
\multicolumn{1}{c}{} & \multicolumn{8}{c}{\textbf{In-Domain}} & \multicolumn{12}{c}{\textbf{Cross-Domain}} \\ 
\cmidrule(r){2-9} \cmidrule(l){10-21}
\multirow{2}{*}{\textbf{Method}} & \multicolumn{4}{c}{\textbf{COCO}}  & \multicolumn{4}{c}{\textbf{Flickr30k}} & \multicolumn{4}{c}{\textbf{COCO $\Rightarrow$ Flickr30k}} & \multicolumn{4}{c}{\textbf{Flickr30k $\Rightarrow$ COCO}} & \multicolumn{4}{c}{\textbf{COCO $\Rightarrow$ NoCaps}} \\
& B@4 &  M & C & S & B@4 &  M & C & S & B@4 &  M & C & S & B@4 &  M & C & S & I & N & O & E \\ 
\midrule
CapDec~\shortcite{capdec} & 26.4 & 25.1 & 91.8 & -    & 17.7 & 20.0 & 39.1 & - & 17.3 & 18.6 & 35.7 &    - & 9.2 & 16.3 & 27.3 &   - & 60.1 & 50.2 & 28.7 & 45.9\\
CapDec$+\text{$\ourmodule$}$ & 27.0 & 25.6 & 94.2 & 18.8 & 18.6 & 20.1 & 44.4 & 14.4 & 15.7 & 18.0 & 35.8 & 11.8 & 9.5 & 16.3 & 30.7 & 9.2 & 60.2 & 50.9 & 33.7 & 47.8\\
\multicolumn{1}{c|}{$\mathrm{\Delta}$} & {\color{ForestGreen}\textbf{0.6}} 
     & {\color{ForestGreen}\textbf{0.5}} 
     & {\color{ForestGreen}\textbf{2.4}} & - 
     & {\color{ForestGreen}\textbf{0.9}} 
     & {\color{ForestGreen}\textbf{0.1}} 
     & {\color{ForestGreen}\textbf{5.3}} & -
          & {\color{red}\textbf{-1.6}} 
     & {\color{red}\textbf{-0.6}} 
     & {\color{ForestGreen}\textbf{0.1}}
     & -
     & {\color{ForestGreen}\textbf{0.3}}
     & {\color{ForestGreen}\textbf{-}} 
     & {\color{ForestGreen}\textbf{3.4}}
     & -
     & {\color{ForestGreen}\textbf{0.1}} 
     & {\color{ForestGreen}\textbf{0.7}} 
     & {\color{ForestGreen}\textbf{5.0}}
     & {\color{ForestGreen}\textbf{1.9}}\\
\midrule
ViECap~\shortcite{viecap} & 27.2 & 24.8 & 92.9 & 18.2 & 21.4 & 20.1 & 47.9 & 13.6 & 17.4 & 18.0 & 38.4 & 11.2 & 12.6 & 19.3 & 54.2 & 12.5 & 61.1 & 64.3 & 65.0 & 66.2 \\
ViECap$+\text{$\ourmodule$}$ & 27.3 & 25.1 & 93.6 & 18.4 & 21.2 & 20.2 & 48.8 & 13.9 & 17.4 & 18.1 & 40.2 & 11.1 & 13.6 & 19.3 & 55.2 & 12.7 & 62.2 & 64.9 & 67.1 & 67.2 \\
\multicolumn{1}{c|}{$\mathrm{\Delta}$} & 
{\color{ForestGreen}\textbf{0.1}} 
     & {\color{ForestGreen}\textbf{0.3}} 
     & {\color{ForestGreen}\textbf{0.7}}
     & {\color{ForestGreen}\textbf{0.2}}
     & {\color{red}\textbf{-0.2}} 
     & {\color{ForestGreen}\textbf{0.1}} 
     & {\color{ForestGreen}\textbf{0.9}}
     & {\color{ForestGreen}\textbf{0.3}}
     & {\color{ForestGreen}\textbf{-}} 
     & {\color{ForestGreen}\textbf{0.1}} 
     & {\color{ForestGreen}\textbf{1.8}}
     & {\color{red}\textbf{-0.1}}
     & {\color{ForestGreen}\textbf{1.0}}
     & {\color{ForestGreen}\textbf{-}} 
     & {\color{ForestGreen}\textbf{1.0}}
     & {\color{ForestGreen}\textbf{0.2}}
     & {\color{ForestGreen}\textbf{1.1}} 
     & {\color{ForestGreen}\textbf{0.6}} 
     & {\color{ForestGreen}\textbf{2.1}}
     & {\color{ForestGreen}\textbf{1.0}}\\
\bottomrule[1pt]
\end{tabular}%
}
\caption{Results of applying the $\ourmodule$ module to the text-only training models: In-domain results on COCO and Flickr30k test sets; cross-domain results on Flickr30k, COCO test sets, and NoCaps validation set. CIDEr score is used for NoCaps evaluation. I is in-domain, N denotes near-domain, O represents out-of-domain, and E indicates entire dataset. Our method consistently improves performance across most metrics in both in-domain and cross-domain settings, regardless of the base models. Notably, a significant performance improvements in CIDEr is observed in cross-domain scenarios when combining $\ourmodule$ module.
}\label{tb:Vipmodule_all_domain}
\end{center}
\end{table*}


\begin{table}[t]
\begin{center}

\setlength{\tabcolsep}{1mm}
\begin{tabular}{l|ccccc}
\toprule[1pt]
\multirow{2}{*}{}    & $\textit{No}$ & \multirow{2}{*}{$\mathcal{N}(0, 1)$} & \multirow{2}{*}{Unif(-1,1)} & Unif(0,1)                     & $\alpha\cdot\omega_{add}$ \\
                      &            $\textit{Noise}$               &                          &                             & $\times\mathcal{N}(0,1)$ & (Ours)                      \\ 
                      \midrule

{CIDEr} & 121.1                     & 121.4                    & 121.9                       & 122.0                         & \textbf{122.9}                 \\
\bottomrule[1pt]
\end{tabular}
\caption{CIDEr scores estimated by adding vectors sampled from different probability distributions to the semantic features in the COCO test dataset. Uniform distribution denotes as Unif. Unif(0,1)$\times\mathcal{N}(0,1)$ represents the method proposed by LinCIR~\cite{lincir}.}\label{tb:noise_type}

\end{center}

\end{table}

During training, the $\ourmodule$ module uses a patch size of $M$=200, and hyperparameter $\alpha$ is set to 5. The model selects three retrieved captions per image (k=3) from the COCO datastore due to the limitation of 77 context length size in CLIP. Captions are retrieved using CLIP ResNet-50x64 and processed with FAISS~\cite{faiss} for fast nearest neighbor search. 
Caption quality is evaluated using the metrics BLEU@4 (B@4)~\cite{bleu}, METEOR (M)~\cite{meteor}, CIDEr (C)~\cite{cider}, and SPICE (S)~\cite{spice}.

\subsection{Main Results}
\noindent\textbf{In-domain.}
We evaluate $\model$ on the COCO, Flickr30k, and NoCaps datasets in \Tref{tb:main}. \Tref{tb:main} shows the evaluation results on the COCO dataset. The upper part of \Tref{tb:main} refers to the performance of large models trained on large datasets. In particular, compared to OSCAR$_\text{Large}$, our model outperforms in B@4 score while using only 4$\%$ of the parameters. Our model, with 5 times fewer parameters than CaMEL, obtains the second-highest CIDEr score among lightweight models. In the COCO dataset, $\model$ exceeds the baseline, SmallCap, and outperforms SmallCap$_\text{Large}$, even though it has significantly fewer parameters (14M vs. 47M). The model also shows strong performance on the Flickr30k dataset.

\begin{table}[t]
    \begin{center}
        
    \setlength{\tabcolsep}{1mm}

\begin{tabular}{l|c|c|c|c|c}
\toprule[1pt]
\textbf{Method}        & \textbf{Enc.} & \textbf{Dec.}         & \textbf{$\ourmodule$}       & \textbf{Ret} & \textbf{CIDEr} \\ \midrule
                       &               & OPT                   & \multirow{4}{*}{$\checkmark$ } & $\times$   & 122.0          \\
$\model$               & ViT           & -125M                 &                    & $\checkmark$    & 122.5 ({\color{ForestGreen}\textbf{0.5 $\uparrow$}})          \\ \cmidrule{3-3} \cmidrule{5-6} 
(Ours)                   & -B/32         & \multirow{2}{*}{XGLM} &                    & $\times$   & 116.8          \\
                       &               &                       &                    & $\checkmark$    & 121.2 ({\color{ForestGreen}\textbf{4.4 $\uparrow$}})          \\ \midrule
\multirow{2}{*}{EVCap} & EVA-          & Vicuna                & $\times$                  & $\checkmark$    & 140.1          \\
                       & CLIP-g        & -13B                  & $\checkmark$                  & $\checkmark$     & 141.3 ({\color{ForestGreen}\textbf{1.2 $\uparrow$}})          \\ \bottomrule[1pt]
\end{tabular}

\caption{Performance improvements in CIDEr scores on COCO test across various decoders using the $\ourmodule$ module and the retrieved text. Ret refers to the usage of the retrieved text.
It also maintains performance improvements when changing the encoder and decoder to larger models.
}\label{tb:model_agnostic}

\end{center}
\end{table}

\noindent\textbf{Cross-domain.}
On the Nocaps validation dataset in the \Tref{tb:main}, it achieves a CIDEr score of 93.8 in in-domain data, surpassing all lightweight models. $\model$ exhibits superior performance in both in-domain and cross-domain, notably outperforming the previous SOTA models by over 3 points across most of metrics in the cross-domain. Additionally, it exceeds the large-scale training model, Oscar$_\text{Large}$, by more than 10 points on in-domain data, indicating further performance in entire data. \Tref{tb:main} shows that $\model$ is highly suitable for zero-shot tasks and real-world scenarios.


\noindent\textbf{Additional vector design.} We add vectors to enable the model to capture more complex data structures when estimating the Gaussian distribution. In the \Tref{tb:noise_type}, we experiment by adding a vector to the semantic features across various design cases. $\textit{No noise}$ is adding nothing. 
We empirically find that our method enhances performance compared to not adding a vector or adding a vector sampled from a different distribution during the sampling process.

\begin{table}[t]
\begin{center}
\setlength{\tabcolsep}{1mm}

\begin{tabular}{l|c|c|c|c|c}
\toprule[1pt]
\textbf{Method}       & \textbf{Enc.}         & \textbf{Dec.} & \textbf{Ret.}       & \textbf{B@4}          & \textbf{CIDEr}         \\ \midrule
\multirow{2}{*}{MAPL} &                       & GPT-J         & \multirow{2}{*}{$\times$} & \multirow{2}{*}{36.5} & \multirow{2}{*}{125.2} \\
                      & ViT                   & 6B            &                    &                       &                        \\ \cmidrule{1-1} \cmidrule{3-6} 
ViPCap                & -L/14                 & GPT2-L        & $\times$                 & 38.9 ({\color{ForestGreen}\textbf{2.4 $\uparrow$}})           & 128.3 ({\color{ForestGreen}\textbf{3.1 $\uparrow$}})            \\
(Ours)                & \multicolumn{1}{l|}{} & 0.7B          & $\checkmark$                 & \textbf{40.5 ({\color{ForestGreen}\textbf{4.0 $\uparrow$}})}         & \textbf{131.0 ({\color{ForestGreen}\textbf{5.8 $\uparrow$}})}         \\ \bottomrule[1pt]
\end{tabular}

    \end{center}    
    \caption{CIDEr and BLEU@4 scores for different types of GPT decoders on the COCO test. Ret refers to the usage of the retrieved text.}
    \label{tb:model_agnostic2}
\end{table}

\subsection{$\ourmodule$ Module Capability}
We explore different model sizes and prompt styles to evaluate the capabilities of our model.

\noindent\textbf{Plug-and-Play manner.} 
In \Tref{tb:Vipmodule_all_domain}, we evaluate the COCO, Flickr30k, and NoCaps test datasets to explicitly demonstrate the visual prompt ability by the $\ourmodule$ module. 
We combine our module with text-only training models. 

As a result, applying the $\ourmodule$ module to CapDec and ViECap~\cite{capdec, viecap} improves performance, demonstrating that the $\ourmodule$ module can function as both a visual prompt and an image feature. The $\ourmodule$ module easily fuses with the vision encoders without modifying the framework. 
CapDec and ViECap with the $\ourmodule$ module result in an average increase of 3.5 points in CIDEr score on cross-domain in the NoCaps dataset. Our method shows the capability of $\ourmodule$ module in zero-shot tasks across real-world scenarios. We do not test with DeCap~\cite{decap} due to its focus on memory efficiency, which does not align with our goals.


\begin{table}[t]
\centering
            \begin{tabular}[t]{lcc}
            \toprule[1pt]
            \multirow{2}{*}{\textbf{Prompt}} & \multicolumn{2}{|c}{\textbf{$\ourmodule$}} \\ 
             & \multicolumn{1}{|c|}{$\times$} & $\checkmark$ \\
            \midrule
            \multicolumn{1}{l|}{``This image shows''} & \multicolumn{1}{l|}{111.1} & 116.0 ({\color{ForestGreen}\textbf{4.9 $\uparrow$}}) \\
            \midrule
            \multicolumn{1}{l|}{Retrieval prompt}  & \multicolumn{1}{l|}{117.3} & 119.9 ({\color{ForestGreen}\textbf{2.6 $\uparrow$}}) \\
            \bottomrule[1pt]
            \end{tabular}
        \caption{
        CIDEr results of the ViP module on COCO val dataset show the potential of our prompt-agnostic model.
        }\label{tb:Vpp_prompt_agnostic}
\end{table}



\begin{table}[t]
\centering
\begin{tabular}{lcccc}
\toprule[1pt]
\multicolumn{1}{c}{\textbf{$PR$}} & \multicolumn{1}{c}{\bm{$\omega$}} & \multicolumn{1}{c|}{\bm{$\alpha$}} & \textbf{B@4} & \textbf{CIDEr} \\ \midrule
$\times$                                & $\checkmark$                               & \multicolumn{1}{c|}{$\times$}          & 36.6         & 120.1      \\  
$\times$                                & \checkmark                               & \multicolumn{1}{c|}{\checkmark}          & 36.9         & 121.1      \\ 
\checkmark                                & $\times$                               & \multicolumn{1}{c|}{$\times$}          & 37.0         & 121.1      \\ 
\checkmark                                & \checkmark                               & \multicolumn{1}{c|}{$\times$}          & 37.3         & 121.9      \\  
\checkmark                                & \checkmark                               & \multicolumn{1}{c|}{\checkmark}          & \textbf{37.7}         & \textbf{122.9}      \\ \bottomrule[1pt]
\end{tabular}
\caption{Experimental results on the COCO test dataset. $PR$ denotes the Patch Retrieval module, $\omega$ is the learnable vector, and $\alpha$ represents the scale factor used in the $\alpha\cdot\omega_{add}$ with sampling Gaussian distribution.
}
\label{tb:tb_components}
\end{table}



\begin{table}[t]
\begin{center}
\begin{tabular}{lcc}
\toprule[1pt]
\multicolumn{1}{l|}{\textbf{FFN}}
                                              & \textbf{B@4} & \textbf{CIDEr} \\ \midrule
\multicolumn{1}{l|}{Sum}                      &  36.4  & 119.1  \\
\multicolumn{1}{l|}{Concat}                   &  36.9 &  119.8 \\
\multicolumn{1}{l|}{MLP$\rightarrow$Sum}    & 36.8  & 119.8  \\
\multicolumn{1}{l|}{MLP$\rightarrow$Concat} & 37.1  & 119.8  \\ 
\midrule
\multicolumn{1}{l|}{Ours} &  \textbf{37.7} & \textbf{122.9}  \\ 
\bottomrule[1pt]
\end{tabular}
\caption{CIDEr and BLEU@4 on COCO test dataset for different Feature Fusion Network (FFN) designs.}
\label{tb:fusion network}
\end{center}
\end{table}

\noindent\textbf{Model-agnostic.}
\Tref{tb:model_agnostic} reveals that combining the $\ourmodule$ module with OPT~\cite{opt} and XGLM~\cite{xglm} as decoders, along with using both $\ourmodule$ and the retrieved text, leads to a notable improvement in performance. This indicates the capability as a model-agnostic and flexible framework. Additionally, similar to SamllCap, combining EVCap~\cite{evcap}, which utilizes retrieval data, with the $\ourmodule$ module enhances performance. This means our approach can be effectively applied to models leveraging retrieval data, as well as to large-scale models such as EVA-CLIP and Vicuna~\cite{evaclip, vicuna2023}.

We cannot apply our module to MAPL~\cite{mapl} due to its limited access to the training code. Hence, in \Tref{tb:model_agnostic2}, 
we evaluate the same encoder with different types of GPT decoders. We use ``This image shows'' as the hard prompt. Notably, despite our decoder having approximately 9 times fewer parameters, it achieves a CIDEr score improvement of over 3.1 points and nearly 6 points when using the retrieved text. Therefore, $\ourmodule$ module presents a consistent performance improvement when combined with SOTA models regardless of model size.

\begin{table}[t]
\begin{center}
\setlength{\tabcolsep}{1mm}
\begin{tabular}{l|ccccccc}
\toprule[1pt]
$M$     & 100 & 150 & 200 & 250 & 300 & 400 & 500 \\ 
\midrule
B@4    &  37.3    &   37.2  &  \textbf{37.7}   &   37.3  &   37.2 & 37.2 &  36.9   \\ \midrule
CIDEr &  121.4   &  121.1   &  \textbf{122.9}   &  121.5   &  121.4   &  121.4 & 121.1  \\ 
\bottomrule[1pt]
\end{tabular}
\end{center}
\caption{CIDEr and BLEU@4 scores on the COCO test data depending on sampling $M$ number of semantic feature.}
\label{tb:m_vs_cider}
\end{table}


\noindent\textbf{Prompt-agnostic.} \Tref{tb:Vpp_prompt_agnostic} compares $\model$ performance with and without the retrieval module. SmallCap scores 111.1 without retrieval and 117.3 with retrieval. $\model$ scores 116.0 without retrieval, using simple prompts like ``This image shows ...'' and 119.9 with retrieval. 
Additionally, in CapDec, ViECap, and EVCap, we observe notable results by leveraging hard prompts such as ``a photo of'' and ``There are $entity_1$, $entity_2$, ...''.
This demonstrates that $\model$ addresses competitive performance even with simple hard prompts and suggests its potential as a flexible visual prompt module applicable to various prompt types. 

\subsection{Ablation Studies}\label{sec:ablation}
We conduct ablation studies to evaluate the effects of various components in our work, including the design of $\ourmodule$ module, additional vector, and feature fusion network strategy.

\noindent\textbf{Effect of $\ourmodule$ module components.} We explore the effect of the patch retrieval module, learnable vector, and scale factor in  \Tref{tb:tb_components}. 
Experimental results show that the patch retrieval module achieves better performance, which can be attributed to its close alignment with visual features.
Adding a learnable vector during sampling improves performance. Scaling this vector with the $\omega$ parameter provides the optimal framework, and all experiments in this paper are conducted with the same scaling value.

\noindent\textbf{Feature Fusion Network design.}\label{sec:ffn_design} In the \Tref{tb:fusion network}, our FFN shows more effectiveness in mitigating the modality gap between image and text semantic features, and generating refined visual features with only a single layer. Compared to our approach, the summation, concatenation, and MLP methods struggle to capture distribution and relational context between modalities.

\noindent\textbf{Effect of $M$ number sampling.}\label{sec:M_num} We extracts semantic features through $M$ number sampling. In \Tref{tb:m_vs_cider}, where $M$ number of sampling ranges from 100 to 500, the highest score is observed at $M = 200$. When there are relatively too many patches ($M > 200$), the local representation $G$ becomes scattered, and leads to no further improvement in performance.


\section{Conclusion}
\label{sec:conclusion}
In this work, we introduce $\model$, a novel approach that generates visual prompts by leveraging semantic information from retrieved text embedding through the $\ourmodule$ module. $\model$ performs well across both in-domain and out-of-domain datasets. The $\ourmodule$ module proposes a plug-and-play method that generates visual prompts based on various models and prompt types. Future work will explore using learnable tokens as visual prompts for better flexibility.


\section{Acknowledgements}

This was supported by the National Research Foundation of Korea (NRF) grant funded by the Korea government (MSIT) (No. RS-2023-00245661).

\bibliography{aaai25}

\end{document}